\documentclass{article}
\usepackage{amsmath, amssymb}
\usepackage[hyperfootnotes=false]{hyperref}
\usepackage{enumitem}
\usepackage{cite}
\usepackage{authblk}
\usepackage{geometry}
\usepackage{subfigure,graphicx}
\usepackage{multirow}
\usepackage{booktabs}
\usepackage{amsmath}
\usepackage{amsfonts}
\usepackage{algorithm}
\usepackage{algpseudocode}
\usepackage{array}
\usepackage{float}
\usepackage{authblk}
\usepackage{subcaption}
\usepackage{ccicons}

\title{\textbf{Opus} \vspace{0.5em}\\ \Large{{\fontseries{bx} A Large Work Model for Complex Workflow Generation}}}

\author{
\vspace{1em}
\begin{minipage}[t]{0.25\textwidth}
    \centering
    \begin{tabular}[t]{c}
        \small{\textbf{Théo Fagnoni}} \\ \vspace{-0.7em}
        \scriptsize{Member of Technical Staff} \\
        \scriptsize{AppliedAI} \\
    \end{tabular}
\end{minipage}%
\begin{minipage}[t]{0.25\textwidth}
    \centering
    \begin{tabular}[t]{c}
        \small{\textbf{Bellinda Mesbah}} \\ \vspace{-0.7em}
        \scriptsize{Member of Technical Staff} \\
        \scriptsize{AppliedAI} \\
    \end{tabular}
\end{minipage}%
\begin{minipage}[t]{0.25\textwidth}
    \centering
    \begin{tabular}[t]{c}
        \small{\textbf{Mahsun Altin}} \\ \vspace{-0.7em}
        \scriptsize{Member of Technical Staff} \\
        \scriptsize{AppliedAI} \\
    \end{tabular}
\end{minipage}%
\begin{minipage}[t]{0.25\textwidth}
    \centering
    \begin{tabular}[t]{c}
        \small{\text{\textbf{Phillip Kingston}}}\\ \vspace{-0.7em}
        \scriptsize{Member of Technical Staff} \\
        \scriptsize{AppliedAI} \\
        \texttt{phillip.kingston@opus.com} \\
    \end{tabular}
\end{minipage}%
\vspace{1em}
}

\date{\normalsize{12 November 2024}}

\begin{document}

\maketitle

\begin{center}
    \ccbyncsa \\
    \vspace{0.3em}
    \footnotesize{This work is licensed under a Creative Commons Attribution-Noncommercial-ShareAlike 4.0 International License (CC BY-NC-SA 4.0)}
\end{center}

\vspace{1em}

\begin{abstract}

\noindent This paper introduces Opus, a novel framework for generating and optimizing Workflows tailored to complex Business Process Outsourcing (BPO) use cases, focusing on cost reduction and quality enhancement while adhering to established industry processes and operational constraints. Our approach generates executable Workflows from Intention, defined as the alignment of Client Input, Client Output, and Process Context. These Workflows are represented as Directed Acyclic Graphs (DAGs), with nodes as Tasks consisting of sequences of executable Instructions, including tools and human expert reviews. We adopt a two-phase methodology: Workflow Generation and Workflow Optimization. In the Generation phase, Workflows are generated using a Large Work Model (LWM) informed by a Work Knowledge Graph (WKG) that encodes domain-specific procedural and operational knowledge. In the Optimization phase, Workflows are transformed into Workflow Graphs (WFGs), where optimal Workflows are determined through path optimization. Our experiments demonstrate that state-of-the-art Large Language Models (LLMs) face challenges in reliably retrieving detailed process data as well as generating industry-compliant workflows. The key contributions of this paper include:

\begin{enumerate}
    \item The integration of a Work Knowledge Graph (WKG) into a Large Work Model (LWM), enabling the generation of context-aware, semantically aligned, structured and auditable Workflows.
    \item A two-phase approach that combines Workflow Generation from Intention with graph-based Workflow Optimization.
    \item Opus Alpha 1 Large and Opus Alpha 1 Small, models that outperform state-of-the-art LLMs by 38\% and 29\% respectively in Workflow Generation for a Medical Coding use case.
\end{enumerate}

\end{abstract}

\section{Introduction}
In the Business Process Outsourcing (BPO) industry, achieving efficiency and consistent quality in work execution is a challenge, particularly when dealing with complex Workflows that transform a Client’s Input into a defined Output under specific constraints such as time and cost. Most BPO work can be modeled as Workflows, where a series of tasks are executed to transform the Input into the Output. The quality of the Output, as well as adherence to constraints, is ultimately judged by the Client.

\vspace{1em}

In this paper, we represent Workflows as Directed Acyclic Graphs (DAGs). Each node is a Task ($t$), each workflow has single entry and exit Tasks. The entry Task takes as input the Client's Input $(I)$, and the exit Task outputs the final Output $(O)$ to the Client. The edges represent the flow of execution of the Tasks. Tasks consist of sequences of executable Instructions, which can range from simple code to predefined tools, as well as human expert reviews at various stages to ensure quality.

\vspace{1em}

A significant challenge in the modernization of BPO lies in the closed and proprietary nature of Workflow process data, which hinders the ability to explicitly define Workflows and Tasks, and therefore achieving the desired levels of complexity and quality in Workflows for specific use cases.

\vspace{1em}

Our goal is to automate and improve BPO processes, reducing costs and enhancing quality, by generating Workflows that can efficiently transform Input into Output. This includes not only generating Tasks and selecting appropriate Instructions but also deciding when human intervention is necessary. Additionally, in this paper, we address the optimization of Workflows post-generation by refining Tasks, edges, Instructions, and human expert reviews to meet time, accuracy, and cost constraints. This approach is adopted in the \textbf{Opus AI Knowledge Work Platform} that allows organizations who would otherwise use BPOs to reduce costs and increase operational flexibility by combining efficient AI Workflows with human expert reviews.

\paragraph{Definitions}

Our solution is based on the following concepts: 

\textbf{Workflow Intention} is defined as the structured alignment of the Client’s Input (I), potential Process Context, and the expected or actual Client's Output (O).

The \textbf{Work Knowledge Graph (WKG)} is our proprietary graph representing Workflows collected from various use cases across industries, encapsulating domain-specific procedural and operational knowledge.

\textbf{Workflow Generation} is defined as the process of leveraging the Intention representation to identify relevant areas of the WKG in order to generate Workflows as sequences of Tasks using a Large Work Model (LWM).

The \textbf{Large Work Model (LWM)} is our fine-tuned model informed by the WKG and the Intention, constrained to transform Input into Output. The generation phase concludes by combining the Workflows generated by the LWM into a graph, the Workflow Graph (WFG).

The \textbf{Workflow Graph (WFG)} is defined as a directed graph representing multiple connected Workflows from Input (I) to Output (O).

\textbf{Workflow Optimization} is defined as solving a path optimization problem on the WFG, where the output Workflow is represented as a path $p^\star$ that minimizes a cost function $\mathcal{C}$.

\textbf{Workflow Execution} is not covered in this article.

\section{Background}
The generation and optimization of workflows through Large Language Models (LLMs) is a multi-faceted challenge that involves a deep understanding of task decomposition, Intention capture, and knowledge injection. This section reviews key advancements in LLM-based workflow generation, focusing on the techniques, specialized models, and frameworks that contribute to more accurate and efficient workflows. We first examine foundational techniques for generating workflows using LLMs, followed by specialized systems that enhance task complexity handling, the role of knowledge graphs, and methods for optimizing workflows. We conclude by evaluating benchmarks and addressing the need for safeguards to assess the performance of these systems.

\paragraph{Large Language Models for Workflow Generation}
Chain of Thought \cite{CoT} is a prompting technique that enhances LLMs' ability to handle complex reasoning queries by generating a series of logical steps. ReAct \cite{ReAct} combines reasoning, action planning, and external interactions to improve interpretability and reduce errors like hallucinations in tasks such as question-answering and fact-verification. The study \cite{Zheng2024} explores how LLMs structure proof steps through in-context learning, improving workflow generation by incorporating learned steps within the input context. Thinking LLMs \cite{ThinkingLLMs} is a training method that equips LLMs with reasoning and planning capabilities through iterative search and preference optimization, using a judge model to evaluate thought candidates and refine responses for complex workflow generation across diverse domains. Specialized systems employ LLMs as \textit{agents} to perform more complex workflows. MegaAgent \cite{MegaAgent} provides an autonomous framework for large-scale multi-agent systems, enabling dynamic agent generation and task management. Gorilla \cite{Gorilla} is a fine-tuned LLaMA-based model that improves LLMs' accuracy in making API calls, a critical function for integrating external data into workflows. The xLAM series \cite{xLAM} offers open-source Large Action Models, ranging from 1B to 8x22B parameters, trained for agent-specific tasks. These models enhance generalizability, making them well-suited for complex workflow generation across diverse domains. 

\vspace{1em}

These studies underscore a fundamental limitation of LLM-based systems: their dependence on training data constrains their ability to retrieve or infer process context, either due to its absence or because of poorly defined Intentions. To enable effective Workflow Generation, three key priorities emerge: accurately capturing user Intentions, incorporating process-specific knowledge, and clearly delineating Workflows, Tasks, and Instructions.

\paragraph{Intention Capture}

Previous work has explored strategies for intention capture in AI-generated workflows. Intention is All You Need \cite{IntentionIsAllYouNeed} highlights the complexities of using generative AI to discern user intentions in natural language, emphasizing the challenge of ``bridging the gap between language space and the physical world", particularly when diverse intentions risk being homogenized by LLMs. Lumos \cite{Lumos}, a multi-modal question-answering system, captures user intent by integrating text and visual inputs, offering a robust framework for understanding intent in multi-modal contexts. Shankar \cite{Shankar2024} introduces DocETL, a declarative framework for unstructured data processing using LLMs, focusing on accuracy and task optimization through an agent-based system that refines document processing workflows through ``rewrite directives" and incorporates an evaluation mechanism to orchestrate task-specific validation. While these works address intention capture through generative AI, multi-modal integration, or agent-based refinement, they often rely on implicit techniques such as prompt engineering. In contrast, we explicitly define an Intention Layer within the Opus system to formalize Intention capture. This layer leverages attention mechanisms to directly model the interplay between Client Inputs, Client Outputs, and Process Context.

\paragraph{Task layer in Workflow Generation}
There is some consensus on the introduction of a task layer as an intermediate process for workflow generation. This layer typically operates at a semantic level, breaking down a workflow into a logical sequence of tasks (or steps, which we refer to as tasks). Chain of Thought \cite{CoT} introduced this approach, emphasizing task decomposition to structure complex workflows. However, this consensus is still nascent as the definition and role of a task vary widely between studies. Many frameworks constrain tasks to be synonymous with LLM agents executing sub-tasks, often incorporating tool usage (like API calls) as part of their capabilities. 
For example, From Words to Actions \cite{FromWordsToActions} explores hierarchical reinforcement learning, where an LLM ``Planner" organizes tasks into language-based sub-goals for high-level planning and low-level execution, with agents utilizing tools to optimize decision-making. Similarly, the Task Decomposition and Agent Generation (TDAG) framework \cite{Wang2024} employs LLMs as agents to decompose tasks, generate sub-agents, and use tools to navigate complex environments. Talker-Reasoner \cite{Christakopoulou2024} introduces a Reasoner agent tasked with multi-step reasoning, planning, and tool usage, reinforcing the trend of viewing tasks as agent-based processes enhanced by tool interactions. Our approach aims at a more generic and flexible definition of a Task: a sequence of Instructions. Instead of limiting tasks to LLM agents, we treat agents as merely one type of Instruction, equal in status to any other Instruction. We define a Task as a sequence of Instructions that can ultimately be synthesized into executable code, enabling a broader and more adaptable framework for Workflow Generation. This approach inherently supports tool usage as part of the Instruction set.

\paragraph{Knowledge Graph-Informed LLM}
Complex workflow generation with LLMs often requires multiple chat iterations with a domain expert to achieve a partial match to the true processes needed to transition from Client Input to Client Output in accordance with industry practice and incumbent human processes. Critical information is absent from LLMs' training data, particularly in the BPO context, where much of the knowledge is either poorly or not documented. Therefore, workflow generation must be informed by external knowledge sources. We introduce our Work Knowledge Graph (WKG) as a structured representation of workflow knowledge. Graphs are a natural fit for representing workflows, and this paper will explore the features of the WKG. A key step in our approach is mapping the Encoded Intention to relevant regions of the WKG to retrieve nodes and edges from the WKG and later drive workflow generation. To achieve this, we leverage established methods for graph embedding and retrieval, primarily attention-based techniques. Node2vec \cite{Grover2016} serves as a foundational framework, learning continuous feature representations for nodes while preserving network neighborhood structures. It enables richer, task-independent representations, proven effective for multi-label classification and link prediction. Coupled with k-nearest neighbors clustering \cite{Kramer2013}, it enables efficient mapping of relevant graph areas to the Encoded Intention. Graph Attention Networks (GATs) \cite{Veličković2017} extend these capabilities by employing self-attention layers, allowing nodes to selectively attend to their neighbors’ features. Subsequent advancements, such as GATv2 \cite{Brody2021}, address the limitations of static attention mechanisms with dynamic attention for improved expressiveness. Techniques like ADGAT \cite{Zhou2023}, which adapt layer depth based on graph sparsity, further optimize graph traversal and representation. Specialized models have demonstrated the applicability of attention-based graph techniques to support workflow generation. Zhang et al. \cite{Zhang2022} developed a GAT-based model for workflow detection in complex industrial environments. Similarly, Graph Recurrent Attention Networks (GRANs) \cite{Liao2019} have shown promise in generating large graphs efficiently by iteratively constructing nodes and edges with attention-driven conditioning.

\vspace{1em}

By combining these graph embedding and retrieval methods, we ensure that Encoded Intentions are mapped accurately to relevant knowledge areas, positioning the WKG as a critical intermediary in the Workflow Generation pipeline. We incorporate the knowledge graph's structure and content directly into the reasoning and decision-making processes of the LLM performing Workflow Generation, creating a tightly coupled, knowledge-informed Workflow Generation process. AGENTiGraph \cite{Zhao2024} demonstrates how combining Knowledge Graphs (KGs) with Large Language Models (LLMs) can improve domain-specific tasks. By employing a multi-agent framework, it dynamically interprets user intent, integrates new knowledge into workflows, and achieves high performance in task classification and execution. Its successful applications in healthcare and legislation highlight the adaptability of this method for addressing complex queries in specialized fields.

\vspace{1em}

Overall, this process significantly improves the precision and contextual relevance of the generated Workflows, ensuring that they align more closely with the intended tasks and domain-specific requirements. To further refine these Workflows and improve their efficiency, we take the direction to focus on optimization after generation. Workflow Optimization involves tuning the generated output, ensuring that they not only meet the desired outcomes, but also operate efficiently within real-world constraints.

\paragraph{Workflow Optimization}

In this paper, we adopt a two-phase approach: Workflow Generation followed by Workflow Optimization. The generation phase operates at a semantic level, where Tasks are instantiated with semantic features designed to fulfill the Workflow's objectives. This phase ensures that Tasks are aligned with the intended Workflow and work knowledge within a semantic space. The subsequent optimization phase is defined as the process of minimizing a cost function derived from a cost model applied to the Instructions, Tasks and the overall Workflow. The objective is to identify an Optimal Path from Input to Output within the set of generated paths. In terms of cost modeling, we enhance the accuracy of cost, time and computational complexity estimations by defining each Task as a sequential implementation of predefined Instructions. Tools such as Tiramisu \cite{tiramisu} and SonarQube are instrumental in this process, providing reliable estimates of execution costs. Guidance Graph Optimization (GGO) \cite{GGO} offers a relevant approach for optimizing pathfinding in environments with multiple interacting agents. The cost is modeled by assigning action costs to transitions along the edges in the graph. It reflects the action costs across paths and agents, aiming to optimize throughput in dynamic multi-agent pathfinding (MAPF) problems, where both the paths and agents evolve over time. 

\vspace{1em}

A notable example of optimization within this framework is GPTSwarm \cite{GPTSwarm}, which introduces a computational graph framework for coordinating multiple LLM-based agents. In this framework, nodes represent task-specific functions (e.g. LLM queries), and edges denote the information flow between them. GPTSwarm automates two primary optimization processes: node-level optimization, which refines LLM prompts for task-specific accuracy, and edge-level optimization, which reconfigures inter-agent connections to enhance collaboration and reduce redundancy. Our approach improves upon GPTSwarm by introducing a concrete cost function to guide Workflow Optimization after task generation. Although GPTSwarm relies on an abstract utility function to indirectly influence optimization, we explicitly define the cost function, enabling us to evaluate and optimize Workflows more effectively. This cost function considers essential factors such as time, resource usage, and task dependencies, to ensure that the final Workflow is efficient. By clearly formalizing the cost model, our approach provides a systematic and interpretable method for optimizing complex Workflows, which sets it apart from the more abstract formulations found in previous work. Once the cost model is established, well-known graph optimization algorithms like Dijkstra's can be effectively employed. This is particularly relevant in our context because the generated Workflow Graph (WFG) is typically solvable by applying a cost model that integrates the three key dimensions of time, resources, and dependencies. 

\vspace{1em}

In the Appendix, we address the application of reinforcement learning (RL) approaches that integrate the cost dimension during task generation — essentially merging the phases of generation and optimization via the reward function. Although this approach holds promise, we observe limitations when attempting to apply it practically to the context of complex workflow generation. Specifically, the challenges of scaling and adapting RL techniques to handle the intricacies and inter-dependencies of large workflows constrain their effectiveness in this domain. These challenges arise from the need for high computational resources and the difficulty in modeling complex, dynamic task dependencies within a reinforcement learning framework, which limits the broader applicability of this method. 

\vspace{1em}
Additionally, we discuss guardrails for scaled production generative systems with human as input, workflow evaluation and benchmarking, as well as some workflow builder systems including ours.

\section{System Architecture}
This section outlines the Workflow Generation and Optimization layers of the Opus system, represented in Figure 1. 

\begin{figure}
\centering
\includegraphics[width=\linewidth]{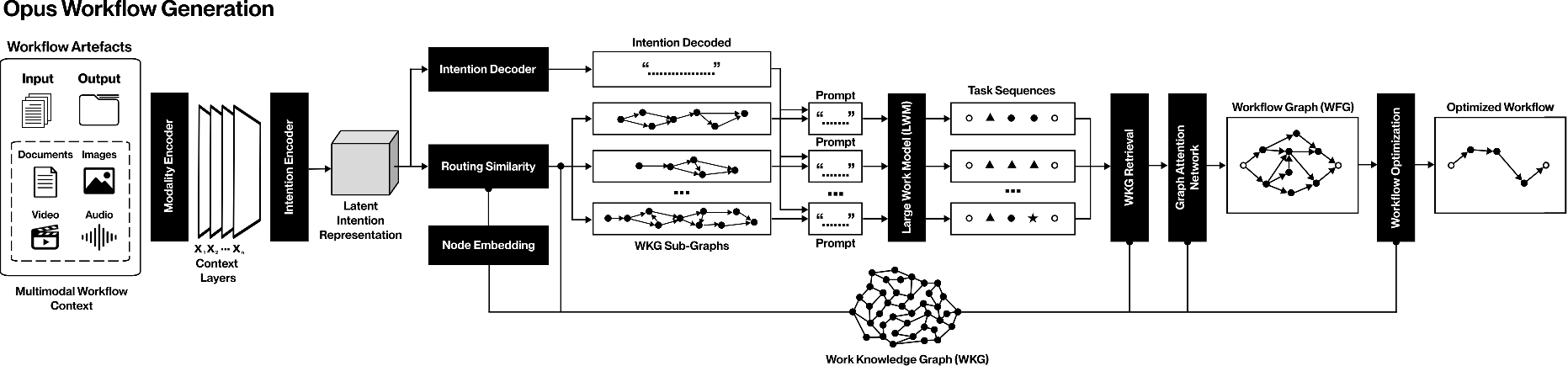}
\caption{Workflow Generation and Optimization}
\label{fig:system_architecture}
\end{figure}

\vspace{1em}

The Opus system consists of four main layers: Intention Encoding, Workflow Graph (WFG) Generation, Workflow Optimization, and Workflow Execution, with the latter not covered in this article. These layers are supported by two key components: the \textbf{Work Knowledge Graph (WKG)}, which organizes domain-specific knowledge, and the \textbf{Large Work Model (LWM)}, designed to interpret and produce workflows. Our approach can be classified as a knowledge graph Retrieval-Augmented Workflow Generation method. The main parts are described in the following sections.

\vspace{1em}

Building on the previously described system input, we encode the Client Input alongside either the Client Output or the Process Context, as at least one of these must be provided. These inputs can theoretically encompass multi-modal data. Each system input is transformed into a unified representation using a dedicated \textbf{Modality Encoder}. 

\vspace{1em}

The input vectors are then merged and encoded through the \textbf{Intention Encoder}, an attention-based neural network that generates a unified Encoded Intention representation. Once the Intention is encoded, two processes occur independently. The first is the \textbf{Intention Decoder}, an attention-based neural network trained to produce a decoded Intention in structured text form. This text describes the Client Input, Client Output, and the required process, ensuring that these three elements are fully embedded and resolved at this stage. The second process identifies relevant areas in the WKG by evaluating the similarity between the Encoded Intention and the embeddings of the graph nodes. This \textbf{Routing Similarity} mechanism ensures that the Encoded Intention is aligned with the appropriate sections of the WKG, setting the foundation for effective Workflow Generation. The retrieved nodes of the WKG are then grouped into sub-graphs, with each sub-graph representing a relevant section of the graph based on the routing similarity. These sub-graphs are converted into textual descriptions by performing a depth-first search (DFS) traversal, during which the semantic features of each node are systematically extracted and translated into coherent text.

\vspace{1em}

For each identified sub-graph, the decoded Intention is incorporated into a prompt alongside the sub-graph’s textual description. This combined prompt serves as input to the LWM, an LLM fine-tuned on Workflow Generation. The model outputs Workflows as a sequence of Tasks, each enriched with semantic features, effectively defining the process required to transition from the Client Input to the Client Output.

\vspace{1em}

Once the Workflows are generated, another retrieval Task is performed on the WKG to identify nodes corresponding to Tasks in the generated Workflows or to locate similar nodes within the WKG in order to retrieve edges and connect Workflows together. For this, a \textbf{Graph Attention Network (GAT)} is employed to represent both the WKG nodes and the nodes from the generated Workflows. These representations are based solely on the semantic features of the nodes and the edges that describe flows of execution. This step is performed until the generated Workflows are integrated into a unified structure forming a Directed Acyclic Graph (DAG) referred to as the \textbf{WFG}.

\vspace{1em}
The final step covered in this paper is \textbf{Workflow Optimization}. This process is formulated as a path optimization problem, determining an optimal sequence of Tasks within the WFG to produce the final Workflow, efficiently transitioning from the Client Input to the Client Output.

\newpage

\section{Workflow Generation and Optimization}
\subsection{Work Knowledge Graph (WKG)}
The WKG is a Directed Graph whose Nodes and Edges are derived from Workflows collected across industries and Opus usage. The following paragraphs describe some of the Node and Edge features, which we leverage for different purposes across the Workflow Generation, Optimization and Execution. 
This paper serves as an illustrative example of our methodology. Only the features leveraged for Workflow Generation will be described, and we omit how we designed and manage the WKG.

\paragraph{Preliminaries}
Let $w$ be a Workflow, we define $i_{w}$ a Workflow Implementation of $w$ into a computer executable script and $\mathcal{I}_w$ the set of Workflow Implementations of $w$. Similarly, $i_{t}$ denotes an implementation of a Task $t$ and $i_{x}$ an implementation of an Instruction $x$. 

\vspace{1em}

\noindent A Workflow Implementation $i_{w}$ is partially composed of a sequence of Task Implementations $(i_{t_., k})_k$. A Task Implementation $i_{t}$ is partially composed of a sequence of Instruction Implementations $(i_{x_., k})_k$, which we will not cover for this version of Workflow Generation and Optimization.

\vspace{1em}

\noindent Each historical Implementation (Workflow, Task, Instruction) is stored along with historical data informing on costs (compute, time, etc.) and success rate which are leveraged for Workflow Optimization.

\vspace{1em}

\noindent We denote by $W$ the set of all Workflows and $\mathcal{I}_W = \bigcup_{w \in W} \mathcal{I}_w$ the set of all historical Workflow Implementations.

\paragraph{Nodes $N_{\text{WKG}}$: Tasks}
In the WKG, a Node is a Task. It has Semantic Features as well as a set of historical Task Implementations. For this version of Workflow Generation, only the Semantic Features of the Tasks are used, which includes Title, Description, Industry, formatted strings of Task Implementations which sequenced Instructions semantic features, etc.

\paragraph{Edges $E_{\text{WKG}}$: Historical Workflow Implementations}
Two Tasks are connected in the WKG if there exists a historical Workflow Implementation where Implementations of these Tasks are consecutive. This establishes the presence of an Edge and its direction. Only this information is leveraged for the Workflow Generation solution described in this paper.

\vspace{1em}
\noindent Considering two Tasks: Task $m$ and Task $n$, let $\mathcal{S}_{t_m, t_n}$ be the set of all consecutive Task Implementations of $t_m$ and $t_n$ across all Workflow Implementations, such that

\begin{equation}
    \mathcal{S}_{t_m, t_n} = \{
    (a, b) \text{ s.t. }
    \exists \hspace{1mm} k\in \mathbb{N}, i_{w} \in \mathcal{I}_W \text{ s.t. } a = i_{t_m, k} \in i_{w} \text{ and } b = i_{t_n, k + 1} \in i_{w}
    \}
\end{equation}

\vspace{1em}
We define $\gamma_{m, n}$ the edge weight between Task $m$ and Task $n$, such that 

\begin{equation}
    \gamma_{m, n} = 1 - e^{-\lambda |\mathcal{S}_{t_m, t_n}|}, \lambda \in ]0,1], \gamma \in [0,1]
\end{equation}
The edge weight between two tasks increases with the extent of their shared historical implementation, reflecting stronger connections for tasks with prior interactions in observed workflows. Edge weights are leveraged at Workflow Optimization.

\subsection{Intention}
The Intention Capture problem involves extracting both process-oriented and result-driven features from system inputs, specifically Client Input, Client Output, and Process Context, to generate effective Workflows, with the Intention Encoder (using attention networks) ensuring that missing components — whether process or output-related— are resolved to align the Workflow with the desired outcome.

\paragraph{Multi-modal Input Pre-processing}
The Opus system supports multiple modalities to capture as much information as possible on the Process Context and the expected Client Output given Client Inputs, i.e. the Intention. Text, images, audio, and video Inputs $(I_m)_m$ are preprocessed independently into a default format: 

\begin{equation}
\hat{I}_m = f^{preprocessing}_{m}(I_m), \quad m \in \{\text{text}, \text{image}, \text{audio}, \text{video} \}
\end{equation}

\paragraph{Multi-modal Input Encoding}
An encoder $ f^{encode}_{m} $ is developed for each modality $m$ in order to get a common vector representation of the preprocessed input $\hat{I}_m$.

\begin{equation}
\Gamma_m = f^{encode}_{m}(\hat{I}_m)
\end{equation}

where $\Gamma_m$ denotes the encoded features of each input of modality $m$ in a common high-dimensional space, enabling cohesive interpretation across modalities in subsequent stages. Text data is encoded through tokenization and attention layers, capturing semantic meaning. Image data goes through OCR for text encoding and convolutional encoding to extract spatial features.

\paragraph{Intention Encoding, Routing and Decoding} 
We define the Intention Capture problem as extracting both process-oriented and result-driven features from the system input to generate effective Workflows. The problem is formalized with three system input components: Client Input (e.g. an input file initiating the Workflow), Client Output (e.g. the expected output file), and Process Context (e.g. training documents defining the Workflow, usually designed to train BPO workers). At least the Client Input, alongside either the Client Output or Process Context, is mandatory for Workflow Generation. We introduce Intention Encoder and Decoder as attention networks, as the problem formulated above closely align with the query-key-value structure \cite{AttentionIsAllYouNeed}. The Client Input acts as the ``query", driving the Workflow Generation by seeking either a defined outcome (Client Output) or procedural details (Process Context). Attention layers compute relevance between query-key pairs to produce the value, which represents the missing information necessary to complete the Intention. If either the Process Context or Client Output is missing, the value provides the missing component. When both Process Context and Client Output are available, the value aligns with the Output, ensuring the Workflow adheres to the desired outcome.

\vspace{1em}
For Intention Encoding, an additional encoder $f^{Intention}$ is developed to create a joint representation $\Gamma$ of all the encoded input modalities, 

\begin{equation}
\Gamma = f^{Intention}(\{\Gamma_m\}_m)
\end{equation}

For Intention Routing, the goal is to utilize the Intention encoding to identify areas of the WKG, which will be leveraged later to generate process compliant Workflows. Each Node from the WKG is text embedded via its semantic features as text inputs are encoded, then compared to the Encoded Intention $\Gamma$ with a similarity function (cosine similarity) and selected with a threshold. At this step, a set $V \subset N_{\text{WKG}}$ of nodes of the WKG is established, highlighting area(s) of the WKG matching the Intention. In the next sections we describe the LWM and how it is used to generate Workflows informed by $V$ and by the Encoded Intention $\Gamma$.

\vspace{1em}

Details on the attention networks used in Intention Routing and Decoding are omitted.

\subsection{Large Work Model (LWM)}
This section outlines the role of the LWM for Workflow Generation.

\paragraph{Incentives for the LWM}
The LWM operates at a semantic level to create Workflows as Task sequences, transforming the Client’s Input into the desired Output. It is guided by two primary incentives for generating Workflows:
\begin{enumerate}
    \item being results-driven: the generated Workflow needs to go from the Client Input to the Client Output, described in the decoded Intention.
    \item being process-driven: the system needs to leverage insights from the WKG to determine how the Workflow should be constructed.
\end{enumerate}

\paragraph{Core functionality and inference}
The LWM is currently a fine-tuned Large Language Model (LLM) trained to generate structured text representing Task sequences. For inference, the Client's Intention is decoded in natural language and each relevant sub-graph of the WKG is converted into text (detailed in next section). The LWM is prompted with these textual descriptions to produce Workflows, one for each sub-graph, detailing each Task’s semantic features including Title, Description, and a structured sequence of Instructions.

\vspace{1em}

The specifics of the LWM model, including its architecture and training details, are omitted.

\subsection{Workflow Graph (WFG)}
In this section we describe the last step of Workflow Generation, the generation of a Workflow Graph (WFG) with the LWM, the WKG, the selected nodes $V$, and the decoded Intention. The goal of the WFG generation is to generate a graph connecting the Client Input (I) to the Client Output (O), from which an Optimal Path can be computed (see next section).

\vspace{1em}

The first step is to compute a set of sub-graphs of the WKG: 
\begin{equation}
    \Theta = \{ \text{SWKG} \}
\end{equation}

We start by splitting $V$ into neighborhoods by computing node embeddings and running a KNN algorithm to get the sets of neighbor nodes.

\begin{equation}
    V = \bigcup N
\end{equation}

For each N, we define SWKG as the minimum spanning tree with edges and nodes from WKG such that 
\begin{equation}
    \forall v \in N, v \in \text{SWKG}
\end{equation}

\vspace{1em}

$\Theta$ being computed, for all SWKG, we generate a sequence of Tasks denoted by $s$, going from Input (I) to Output (O) with the LWM, prompted with:
\begin{enumerate}
    \item decoded Intention as language
    \item formatted text string of a Depth First Search on the SWKG
\end{enumerate}

\vspace{1em}
Each generated sequence is composed of nodes originating from the LWM’s training data, enabling flexibility to go from Input to Output while being process-driven by Work Knowledge injection. Each generated sequence $s = (v_i)_i$ is such that
\begin{equation}
     \{v_i\}_i = \{v_i \in s \text{ and } v_i \in \text{SWKG}\}_i \sqcup \{v_i \in s \text{ and } v_i \notin \text{SWKG}\}_i
\end{equation}

$\{v_i \in s \text{ and } v_i \notin \text{SWKG}\}$ comes from the LWM training data.

\vspace{1em}

Let $S$ be the set of all generated sequences going from Input to Output, we finally build the WFG from $S$, denoted by
\begin{equation}
     \text{WFG} = (V_{\text{WFG}}, E_{\text{WFG}})
\end{equation}

The first step is to convert each sequence $s \in S$ as a Directed Acyclic Graph (effectively as a Workflow), denoted by $\text{dag}_s$, by analyzing dependencies and compatibilities between elements of $s$:
\[
\text{dag}_s: (V_{\text{dag}_s}, E_{\text{dag}_s}), V_{\text{dag}_s} = \{v_i\}_i
\]

We then connect the elements of $D_S = \{\text{dag}_s\}_{s \in S}$ by picking edges from WKG, effectively building $V_{\text{WFG}}$ and $E_{\text{WFG}}$ following algorithm 1.
\begin{algorithm}
\caption{WFG from WKG Edges}
\begin{algorithmic}
\State \textbf{Input:} WKG, $D_S$,
\State \textbf{Output:} WFG
\State
\State \textbf{Initialization}
\State $V_{\text{WFG}} = \bigcup_{s \in S} \{v \text{ for } v \in V_{\text{dag}_s}\}$
\State $E_{\text{WFG}} = \bigcup_{s \in S} \{e \text{ for } e \in E_{\text{dag}_s}\}$
\State
\For{$\text{dag}_s \in D_S$}
    \For {$v \in V_{\text{dag}_s} \cap V_{\text{WKG}}$} 
        \State \(E_{\text{WFG}} = E_{\text{WFG}} \cup \{ e = (v, x) \text{ for } x \in \bigcup_{\text{d} \in D_S \backslash \text{dag}_s} \{v  \text{ for } v \in V_{\text{d}} \cap V_{\text{WKG}}\} \text{ if } e \in E_{\text{WKG}} \}\)
    \EndFor
\EndFor
\end{algorithmic}
\end{algorithm}

\vspace{1em}
\newpage

The last step, required if the WFG is not weakly connected, optional otherwise (to improve density or other metrics), is to recommend nodes and edges from the WKG to be added to it. To do so, we trained a Graph Attention Network (GAT) on the WKG, denoted by $\text{GAT}_{\text{WKG}}$. We define by the same name, for simplicity, the function that gets the set of nodes $V_{\text{WFG}} \cap V_{\text{WKG}}$ and extend this set with nodes from $V_{\text{WKG}} \backslash V_{\text{WFG}}$ that are connected to at least one node in $V_{\text{WFG}}$ and whose embeddings are close enough to this node by a threshold $\alpha$ (by cosine similarity). The function $\text{GAT}_{\text{WKG}}$ takes as input WKG and WFG, and returns nodes and edges, $N_{\text{GAT}_{\text{WKG}}}$ and $E_{\text{GAT}_{\text{WKG}}}$, to be added to the Workflow Graph, as described in algorithm 2.

\begin{algorithm}
\caption{Enhancing WFG with Attention-based Graph Machine Learning }
\label{algo:2}
\begin{algorithmic}
\State \textbf{Input:} WKG, WFG
\State \textbf{Output:} Enhanced WFG
\State
\State \textbf{Initialization:} Set threshold similarity \( \alpha = 1 \)
\While{$(V_{\text{WFG}}, E_{\text{WFG}})$ is not weakly connected}
    \State \(\alpha \gets \alpha - \Delta \alpha\)
    \State \( N_{\text{GAT}_{\text{WKG}}}, E_{\text{GAT}_{\text{WKG}}} = \text{GAT}_{\text{WKG}}(\text{WKG}, \text{WFG}, \alpha) \)
    \State \( E_{\text{WFG}} \gets E_{\text{WFG}} \cup E_{\text{GAT}_{\text{WKG}}} \)
    \State \( V_{\text{WFG}} \gets V_{\text{WFG}} \cup N_{\text{GAT}_{\text{WKG}}} \)
\EndWhile
\State \Return WFG
\end{algorithmic}
\end{algorithm}

\subsection{Workflow Optimization}
The final step of the Opus system presented in this paper is the Workflow Optimization, which is effectively picking a Directed Acyclic Graph, or a sequence of Tasks to simplify, going from Input to Output, from the WFG. It can be expressed as an optimization problem, 

\begin{equation}
p^* = \arg\min_{p \subset \text{WFG}} C(p)
\end{equation}

where \( C(p) \) is a cost function associated with each potential path \( p \) within the WFG, accounting for factors such as computational resources, processing time, and specific model usage costs. Though we will not expand on the cost model in this paper, we adopt a linear model over these three dimensions:

\begin{equation}
p^* = \arg\min_{p} \left( \alpha \cdot C_{compute}(p) + \beta \cdot C_{time}(p) + \gamma \cdot C_{model}(p) \right)
\end{equation}

where \( C_{compute}(p) \) represents the computational cost, \( C_{time}(p) \) denotes the temporal cost, \( C_{model}(p) \) reflects the model usage cost, and \( \alpha \), \( \beta \), and \( \gamma \) are weight factors that prioritize the relative importance of each cost component. In this version, we further model the path cost $C(p)$ as the sum of the costs of all tasks $t$ within the path $p$: $C(p) = \sum_{t \in p}C(t)$, following the previously described cost model. Consequently, we employ a modified Dijkstra algorithm to determine an efficient path from the Input node to the Output node, adapting the approach to account for node-associated costs rather than edge-associated costs.

\section{Empirical results: an application to Medical Coding}

\subsection{CPT\textsuperscript{\textregistered} Evaluation / Management Medical Coding}
CPT\textsuperscript{\textregistered} Evaluation / Management (E/M) is a widely adopted and semantically complex work in hospital Medical Coding. It involves numerous specific definitions of entities that must be identified within multi-modal inputs. This process is predominantly carried out by medical coders, who adhere to established guidelines such as those issued by the American Medical Association (AMA). However, hospitals often implement customized variations of these guidelines, introducing additional intricacies that are difficult to formalize and capture. In collaboration with hospital Healthpoint Abu Dhabi, we mapped outpatient Medical Coding Workflows and best practices, as performed on thousands of cases every day. Medical Coding occurs at the level of a medical encounter, defined as an interaction where a patient visits the hospital to see a doctor. Documentation related to an outpatient encounter typically includes:
\begin{itemize}
    \item[—] Consultation Notes: the primary document, detailing the physician's notes and description of the encounter.
    \item[—] Order List: a record of tests or medications prescribed during the encounter.
    \item[—] Reports: ancillary documents such as MRI or X-Ray results reviewed during the encounter.
\end{itemize}

Medical coders perform three primary tasks on these documents:
\begin{itemize}
    \item[—] Data Points: identifying specific factual entities in the input documents.
    \item[—] Problem Points: highlighting clinically significant issues.
    \item[—] Level of Risk: assessing the overall risk associated with the encounter.
\end{itemize}

The E/M code is deterministically derived from classifications made in these three tasks. Each task requires semantic searches across the input documents to identify and count entities, applying a series of logical operations to synthesize the results. These operations involve numerous non-trivial conditions and intricate rules. In the context of our Workflow-based approach, each of these three tasks is represented as a succession of Tasks within the overall Workflow. In the Appendix we provide a high-level schema that outlines the reference Workflow used in our experiments, alongside a simplified semantic description of the Tasks involved.

\subsection{Experiments}
We conducted experiments comparing two versions of the Opus system, appliedai-opus-1alpha-large and appliedai-opus-1alpha-small, against state-of-the-art LLMs openai-o1-preview-2024-09-12, openai-gpt-4o-2024-08-06, google-gemini-1.5-pro, google-gemini-1.5-flash, and anthropic-claude-3.5-sonnet. Opus Alpha 1 Large version is supported by larger WKG and LWM  (70B versus 8B in the Small version). 

\vspace{1em}

The evaluation focuses on two core metric classes: Semantic Fidelity and Structural Fidelity. To establish a reliable benchmark, senior professional medical coders manually created a ground truth reference Workflow in the same Task sequence format as the Workflow Generation outputs (described in the Appendix). Each model was prompted to generate a Workflow in the form of a sequence of Tasks, with each Task defined by a Title and a Description. The Workflows were evaluated against the 2021 American Medical Association (AMA) guidelines for CPT\textsuperscript{\textregistered} Evaluation / Management Medical Coding. The Opus system was provided the same input prompt as Process Context as well as sample fully redacted medical encounter records including consultation notes, prescription drug orders, orders of tests and lab results. To ensure robustness, results were averaged across ten (10) trials for each model using the same input. A third-party LLM (GPT-4o) was prompted to evaluate the Tasks of the generated Workflows against the Tasks of the reference Workflow. This evaluation process is binary: for each generated Task in a sequence, it was determined whether it was semantically aligned with a Task in the reference Workflow. Importantly, each reference Task could be allocated to only one generated Task within a sequence.

\paragraph{Coverage Ratio (Semantic)} 
The coverage ratio is defined as the proportion of Tasks in the generated Workflow that can be exclusively matched to a corresponding ground-truth Task in the reference sequence. This ratio quantifies the extent to which the generated Workflow aligns with the content of the predefined reference.

\paragraph{Kendall's Tau (Structural)}
For Structural Fidelity we use the Kendall's Tau metric to assess how well the order of the matched Tasks aligns with the order in the reference sequence. The Kendall's Tau coefficient measures the correlation between the ordering of Tasks in the generated Workflow and the reference sequence, with higher values indicating better alignment. To penalize for Tasks that did not match the reference sequence, we multiplied the Kendall's Tau score by the coverage ratio. This combined metric provides a more comprehensive evaluation of the Workflow’s structural accuracy and alignment with the reference sequence.

\paragraph{Dynamic Time Warping (Structural)} 
Similarly, we employed Dynamic Time Warping (DTW) to assess the alignment of Task sequences in the generated Workflows with the reference sequence. DTW measures the order similarity between two sequences by calculating the optimal sequential match between them, allowing for shifts and distortions. In the context of Workflow Generation, DTW compares the alignment of Task order across different sequences, identifying how well the Tasks in the generated Workflow correspond to the reference sequence despite potential differences in orders. We normalized and multiply it by the coverage ratio as for Kendall's Tau.

\paragraph{BLEU Score (Semantic)}
We use the BLEU score to evaluate the quality of generated Tasks by comparing them to the corresponding Tasks in the reference Workflow. For each Task in the generated Workflow, we calculate the BLEU score with its matched Task from the reference sequence, considering 0 for non-matched Tasks, and then average the scores across all Tasks. This metric allows us to assess the semantic alignment of the generated Tasks with the reference Tasks in terms.

\paragraph{Cosine Similarity (Semantic)}
We use cosine similarity as a metric to evaluate the semantic alignment of the generated Tasks with the reference Tasks. For each Task in the generated Workflow, we compute the cosine similarity between its semantic embedding the one of its matched Task from the reference sequence, considering 0 for non-matched Tasks, and then average the cosine similarity scores across all Tasks.

\subsection{Results}
The results are summarized in the following table, with models ranked based on their overall performance across the metrics, computed as the total covered area of the pentagon formed with the five metric values and equal angles.

\begin{table}[H]
\centering
\renewcommand{\arraystretch}{1.5}
\caption{Results}
\label{tab:results_blend}
\resizebox{\textwidth}{!}{%
\begin{tabular}{l|ccccc}
\hline
\textbf{Model} &
\textbf{Coverage Ratio} &
\textbf{Kendall’s Tau} &
\textbf{Dynamic Time Warping} &
\textbf{Cosine Similarity} &
\textbf{BLEU Score} \\
\hline
\hline
\textbf{appliedai-opus-1alpha-large} &
\text{0.721} &
\textbf{0.498} &
\text{0.715} &
\text{0.485} &
\textbf{0.361} \\
\hline
\textbf{appliedai-opus-1alpha-small} &
\textbf{0.746} &
\text{0.083} &
\textbf{0.719} &
\textbf{0.546} &
\text{0.235} \\
\hline
\text{anthropic-claude-3.5-sonnet} &
\text{0.254} &
\text{0.058} &
\text{0.250} &
\text{0.283} &
\text{0.017} \\
\hline
\text{openai-o1-preview-2024-09-12} &
\text{0.271} &
\text{0.056} &
\text{0.255} &
\text{0.202} &
\text{0.001} \\
\hline
\text{openai-gpt-4o-2024-08-06} &
\text{0.208} &
\text{0.005} &
\text{0.195} &
\text{0.264} &
\text{0.005} \\
\hline
\text{google-gemini-1.5-flash} &
\text{0.083} &
\text{0.202} &
\text{0.078} &
\text{0.209} &
\text{0.012} \\
\hline
\text{google-gemini-1.5-pro} & 
\text{0.108} &
\text{0.102} &
\text{0.102} &
\text{0.142} &
\text{0.011} \\
\hline
\end{tabular}%
}
\end{table}

Opus Alpha 1 Large was the best overall performer followed by Opus Alpha 1 Small and Anthropic Claude 3.5 Sonnet. On average across all metrics evaluated, Opus Alpha 1 Large and Opus Alpha 1 Small outperformed Anthropic Claude 3.5 Sonnet by 38\% and 29\% respectively.
\begin{figure}[H]
    \begin{center}
        \begin{tabular}{cc} 
            {\tiny
            \includegraphics[width=0.35\textwidth]{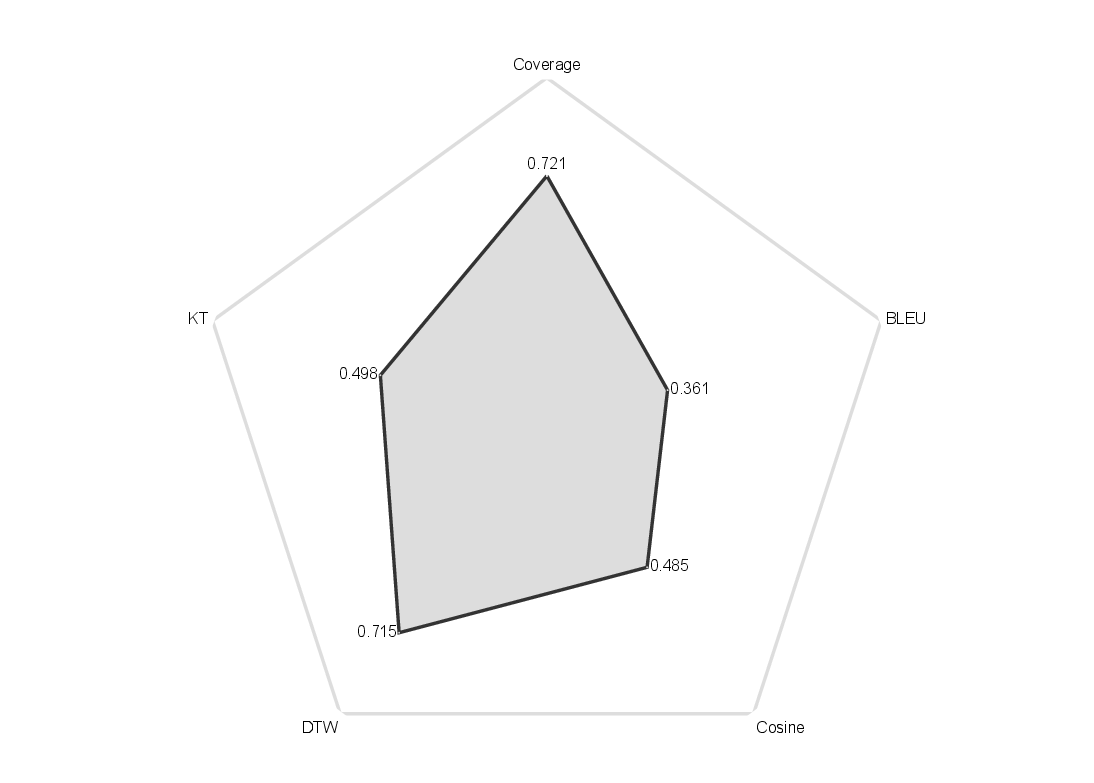}
            } & 
            {\tiny
            \includegraphics[width=0.35\textwidth]{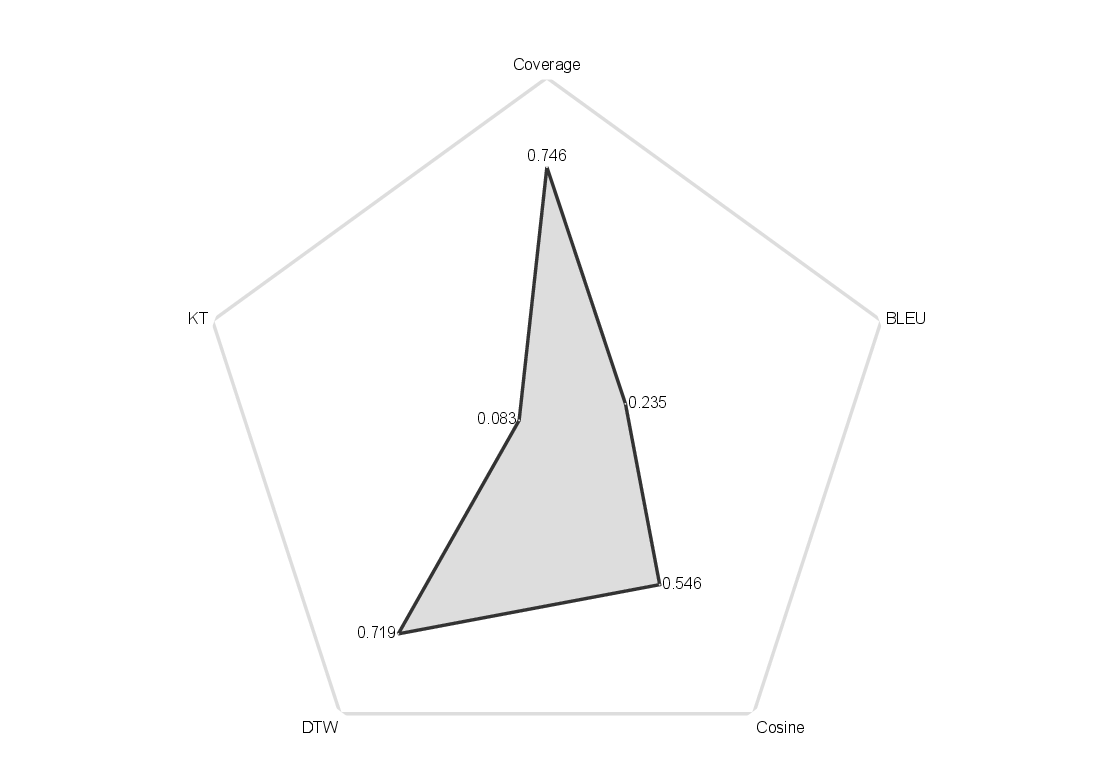} 
            } \\
            Opus Alpha 1 Large & Opus Alpha 1 Small \vspace{0.5em}\\
            {\tiny \includegraphics[width=0.35\textwidth]{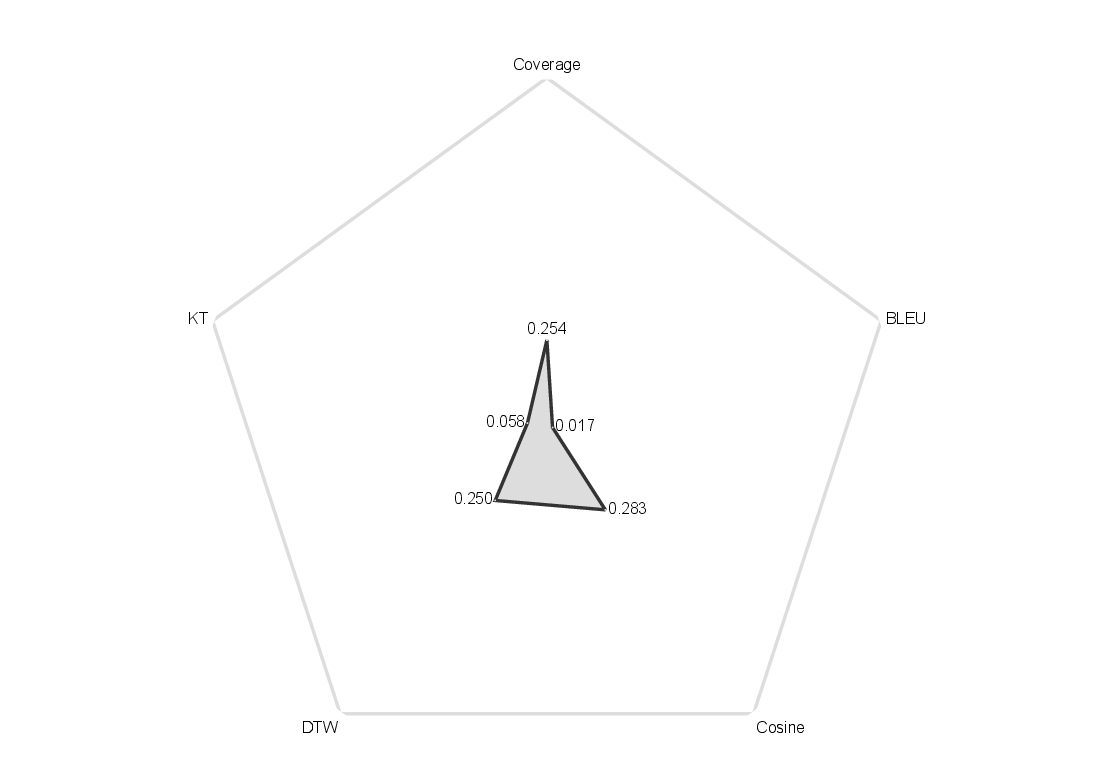} 
            } &
            {\tiny
            \includegraphics[width=0.35\textwidth]{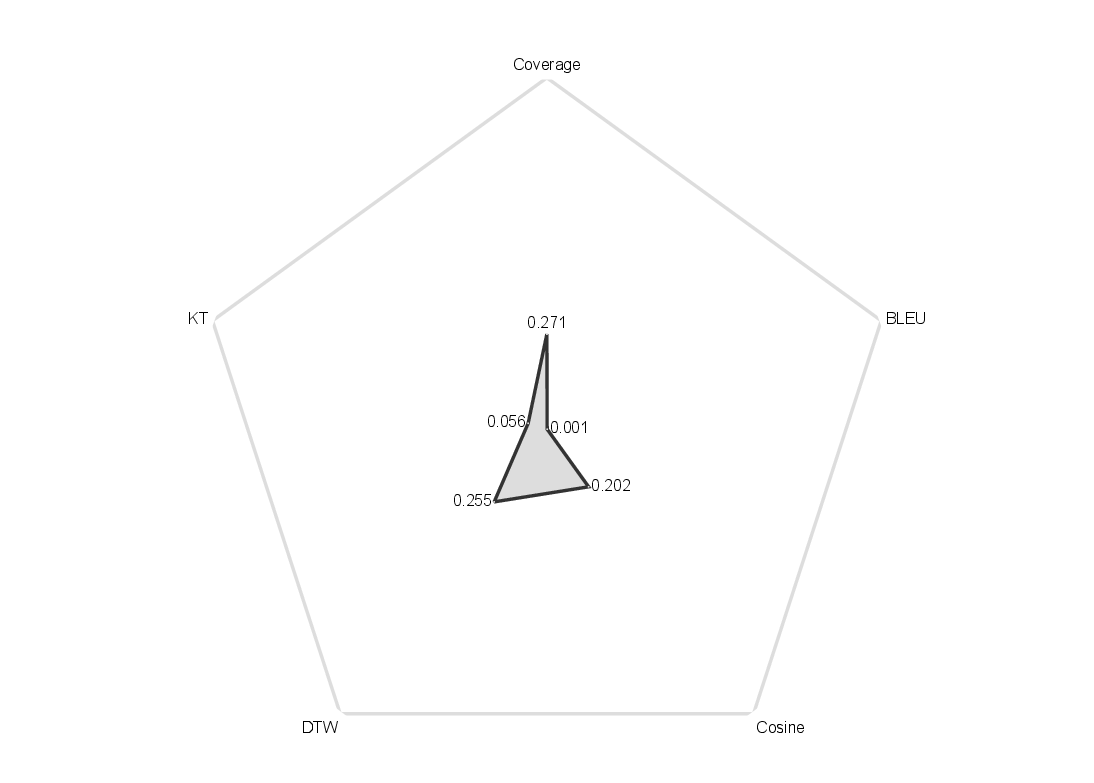}
            } \\
            Anthropic Claude 3.5 Sonnet & OpenAI o1-preview \\
        \end{tabular}
    \end{center}
    \caption{Comparison of Opus Alpha 1 Large and Opus Alpha 1 Small to the top performing state-of-the-art LLMs across the metrics on the Medical Coding use case.}
    \label{fig:bestperformers}
\end{figure}

\paragraph{Coverage Ratio as a Baseline Metric}

The coverage ratio, which is defined as the Task-level semantic alignment between the generated and reference Workflows, serves as the foundational metric for evaluating model quality. The appliedai-opus-1alpha-large and appliedai-opus-1alpha-small models achieve coverage ratios of 0.721 and 0.746, respectively, significantly outperforming other models. These results highlight their superior ability to capture and align Workflow Tasks semantically. We attribute this superior performance to two key innovations: the injection of the WKG, which provides domain-specific procedural knowledge, and the explicit capture of Intention, which ensures alignment of Client Input, Output, and Process Context. Together, these enhancements enable the LWM to generate Workflows with significantly higher semantic alignment and Task-level accuracy. In contrast, industry-standard models, such as anthropic-claude-3.5-sonnet (coverage ratio of 0.254) and openai-gpt-4o-2024-08-06 (coverage ratio of 0.208), perform at substantially lower levels, underscoring the critical role of WKG injection and Intention capture in achieving state-of-the-art results.

\paragraph{Variability between Opus Alpha 1 Large and Opus Alpha 1 Small}

Interestingly, the small variant of the Opus system achieves a higher coverage ratio (0.746) compared to its larger counterpart (0.721), but the larger model demonstrates better performance on certain metrics, such as Kendall’s Tau (0.498 vs. 0.083) and BLEU Score (0.361 vs. 0.235). This indicates that while the smaller model excels in Task coverage — possibly due to reduced noise in the process of identifying areas of a smaller WKG — the larger system generates Workflows with better sequential coherence and linguistic similarity to the reference. This holds true only when the knowledge relevant to the use case is contained within the smaller WKG. Despite these discrepancies, both Opus models outperform the other models by at least a factor of two across most metrics, underscoring their dominance in generating accurate Workflows.

\paragraph{Limitations in Existing Models}
The results reveal that state-of-the-art models, such as openai-gpt-4o-2024-08-06 and google-gemini-1.5-pro, fail to deliver the complexity of Workflows adequately. For example, their coverage ratios are as low as 0.208 and 0.108, respectively, and their BLEU scores, which indicate linguistic alignment, are negligible. These findings suggest that without significant data injection tailored to Workflow Generation, these models struggle to represent the intricate structures required for practical Workflow applications. They are unable to optimize over a Workflow path nor evaluate an unseen Workflow that was not part of their training data.

\vspace{1em}

Results show that standard RAG approaches and state-of-the-art LLMs fail to meet the complexity and precision needed for unambiguous Workflow Generation. These models lack mechanisms for capturing user Intentions and frameworks for storing and leveraging domain-specific knowledge absent from training data. Our integration of Intention Capture and Workflow Knowledge Graph (WKG) to a Large Work Model (LWM) overcomes these limitations: Intention Capture ensures a deep understanding of user inputs, while the WKG provides a scalable, structured repository of domain knowledge. We achieve the generation of detailed, rigorously structured, and logically composed workflows, as demonstrated by superior performance in our use case evaluations. By explicitly defining Instructions, Tasks, and Workflow Layers, our LWM trained in this framework can generate Workflows that address both content gaps and structural deficiencies. While improved prompting may offer minor gains, the challenges of Intention capture, knowledge storage and retrieval, and Workflow composition are only effectively addressed by Opus.

\newpage

\section{Conclusion}
In this article, we presented the Workflow Generation framework of Opus, designed to address the challenges of generating and optimizing complex, industry-level Workflows. Central to our approach is the adoption of a unified framework for representing Workflows, Tasks, and Instructions. A key contribution of our work is the introduction of the concept of Workflow Intention, defined as a three-dimensional alignment of Client Input, Client Output, and Process Context. We proposed a novel methodology for generating Workflows as Directed Acyclic Graphs (DAGs), establishing the importance of Workflows being both process-driven and results-driven. This is achieved by leveraging the concept of Intention, along with the Work Knowledge Graph (WKG), which encapsulates domain-specific procedural knowledge. These components collectively feed the Large Work Model (LWM), enabling it to generate context-aware, semantically aligned, structured and auditable Workflows that can be used in regulated industries. Opus Alpha 1 Large and Opus Alpha 1 Small outperformed state-of-the-art LLMs by 38\% and 29\% respectively in Workflow Generation on a real-world use case in Medical Coding.

\section{Acknowledgments}
The authors would like to extend their gratitude to \textbf{Healthpoint Abu Dhabi, United Arab Emirates}, part of the M42 technology and healthcare group comprising 450 facilities across 26 counties for detailed Workflow design, process and optimization data.

\vspace{1em}

The authors would like to thank \textbf{SRH University of Applied Sciences Heidelberg, Germany} for their oversight of results, evaluation, methodology and analysis.

\vspace{1em}

The authors acknowledge the work and global impact of the \textbf{American Medical Association (AMA)}'s \textbf{Current Procedural Terminology (CPT\textsuperscript{\textregistered}) Evaluation / Management} coding system.

\newpage

\bibliographystyle{plain}

\begin{thebibliography}{9}

\bibitem{CoT}
Wei, J., Wang, X., Schuurmans, D., Bosma, M., Ichter, B., Xia, F., Chi, E., Le, Q., and Zhou, D. (2022). Chain-of-Thought Prompting Elicits Reasoning in Large Language Models. In \emph{Advances in Neural Information Processing Systems}, 35, 24824-24837.

\bibitem{ReAct}
Yao, S., Zhao, J., Yu, D., Du, N., Shafran, I., Narasimhan, K., and Cao, Y. (2022). ReAct: Synergizing Reasoning and Acting in Language Models.

\bibitem{Zheng2024}
Zheng, Z., Malon, C., Min, M. R., and Zhu, X. (2024). Exploring the Role of Reasoning Structures for Constructing Proofs in Multi-Step Natural Language Reasoning with Large Language Models.

\bibitem{ThinkingLLMs}
Wu, T., Lan, J., Yuan, W., Jiao, J., Weston, J., and Sukhbaatar, S. (2024). Thinking LLMs: General Instruction Following with Thought Generation.

\bibitem{MegaAgent}
Wang, Q., Wang, T., Li, Q., Liang, J., and He, B. (2024). MegaAgent: A Practical Framework for Autonomous Cooperation in Large-Scale LLM Agent Systems.

\bibitem{Gorilla}
Patil, S. G., Zhang, T., Wang, X., and Gonzalez, J. E. (2023). Gorilla: Large Language Model Connected with Massive APIs.

\bibitem{xLAM}
Zhang, J., Lan, T., Zhu, M., Liu, Z., Hoang, T., Kokane, S.,  Yao, W., Tan, J., Prabhakar, A., Chen, H., Liu, Z., Feng, Y., Awalgaonkar, T., Murthy, R., Hu, E., Chen, Z., Xu, R., Niebles, J. C., Heinecke, S., Wang, H., Savarese, S., and Xiong, C. (2024). xLAM: A Family of Large Action Models to Empower AI Agent Systems.

\bibitem{IntentionIsAllYouNeed}
Sarkar, A. (2024). Intention Is All You Need. In \emph{Proceedings of the 35th Annual Conference of the Psychology of Programming Interest Group (PPIG 2024)}.

\bibitem{Lumos}
Yin, D., Brahman, F., Ravichander, A., Chandu, K., Chang, K. W., Choi, Y., and Lin, B. Y. (2023). Lumos: Learning Agents with Unified Data, Modular Design, and Open-Source LLMs.

\bibitem{Shankar2024}
Shankar, S., Parameswaran, A. G., and Wu, E. (2024). DocETL: Agentic Query Rewriting and Evaluation for Complex Document Processing.

\bibitem{FromWordsToActions}
He, J., Chen, S., Zhang, F., and Yang, Z. (2024). From Words to Actions: Unveiling the Theoretical Underpinnings of LLM-Driven Autonomous Systems.

\bibitem{Wang2024}
Wang, Y., Wu, Z., Yao, J., and Su, J. (2024). TDAG: A Multi-Agent Framework based on Dynamic Task Decomposition and Agent Generation.

\bibitem{Christakopoulou2024}
Christakopoulou, K., Mourad, S., and Matarić, M. (2024). Agents Thinking Fast and Slow: A Talker-Reasoner Architecture.

\bibitem{Grover2016}
Grover, A., and Leskovec, J. (2016). node2vec: Scalable Feature Learning for Networks. In \emph{Proceedings of the 22nd ACM SIGKDD International Conference on Knowledge Discovery and Data Mining}, 855-864.

\bibitem{Kramer2013}
Kramer, O. (2013). K-Nearest Neighbors. In \emph{Dimensionality Reduction with Unsupervised Nearest Neighbors}, 13-23. Springer.

\bibitem{Veličković2017}
Veličković, P., Cucurull, G., Casanova, A., Romero, A., Lio, P., and Bengio, Y. (2018). Graph Attention Networks. In \emph{International Conference on Learning Representations (ICLR)}.

\bibitem{Brody2021}
Brody, S., Alon, U., and Yahav, E. (2021). How Attentive are Graph Attention Networks? In \emph{International Conference on Learning Representations (ICLR)}.

\bibitem{Zhou2023}
Zhou, J., Du, Y., Zhang, R., and Zhang, R. (2023). Adaptive Depth Graph Attention Networks.

\bibitem{Zhang2022}
Zhang, M., Hu, H., Li, Z., and Chen, J. (2022). Proposal-Based Graph Attention Networks for Workflow Detection. \emph{Neural Processing Letters}, 54(1), 101-123.

\bibitem{Liao2019}
Liao, R., Li, Y., Song, Y., Wang, S., Nash, C., Hamilton, W., Duvenaud, D., Urtasun, R., and Zemel, R. (2019). Efficient Graph Generation with Graph Recurrent Attention Networks. In \emph{Advances in Neural Information Processing Systems (NeurIPS)}, 32.

\bibitem{Zhao2024}
Zhao, X., Blum, M., Yang, R., Yang, B., Marquez-Carpintero, L., Pina-Navarro, M., Wang, T., Li, X., Li, H., Fu, Y., Wang, R., Zhang, J., and Li, I. (2024). AGENTiGraph: An Interactive Knowledge Graph Platform for LLM-based Chatbots Utilizing Private Data.

\bibitem{tiramisu}
Baghdadi, R., Merouani, M., Leghettas, M., Abdous, K., Arbaoui, T., Benatchba, K., and Amarasinghe, S. (2021). A Deep Learning Based Cost Model for Automatic Code Optimization. In \emph{Proceedings of the 2021 International Symposium on Code Generation and Optimization (CGO)}. IEEE.

\bibitem{GGO}
Zhang, Y., Jiang, H., Bhatt, V., Nikolaidis, S., and Li, J. (2024). Guidance Graph Optimization for Lifelong Multi-Agent Path Finding.

\bibitem{GPTSwarm}
Zhuge, M., Wang, W., Kirsch, L., Faccio, F., Khizbullin, D., and Schmidhuber, J. (2024). Language Agents as Optimizable Graphs.

\bibitem{AttentionIsAllYouNeed}
Vaswani, A. (2017). Attention is All you Need. Advances in Neural Information Processing Systems. In \emph{Advances in Neural Information Processing Systems (NeurIPS)}, 30, 5998–6008.

\bibitem{AgentAsAJudge}
Zhuge, M., Zhao, C., Ashley, D., Wang, W., Khizbullin, D., Xiong, Y., Liu, Z., Chang, E., Krishnamoorthi, R., Tian, Y., Shi, Y., Chandra, V., and Schmidhuber, J. (2024). Agent-as-a-Judge: Evaluate Agents with Agents.

\bibitem{Triantafyllou2024}
Triantafyllou, S., Sukovic, A., Zolfimoselo, Y., and Radanovic, G. (2024). Counterfactual Effect Decomposition in Multi-Agent Sequential Decision Making.

\bibitem{Chen2024}
Chen, W., Yuan, J., Qian, C., Yang, C., Liu, Z., and Sun, M. (2024). Optima: Optimizing Effectiveness and Efficiency for LLM-Based Multi-Agent System.

\bibitem{Lee2024}
Lee, D., and Tiwari, M. (2024). Prompt Infection: LLM-to-LLM Prompt Injection within Multi-Agent Systems.

\bibitem{Lermen2024}
Lermen, S., Dziemian, M., and Pimpale, G. (2024). Applying Refusal-Vector Ablation to Llama 3.1 70B Agents.

\bibitem{ChainBuddy}
Zhang, J., and Arawjo, I. (2024). ChainBuddy: An AI Agent System for Generating LLM Pipelines.

\bibitem{Ma2024}
Ma, C., Zhang, J., Zhu, Z., Yang, C., Yang, Y., Jin, Y., Lan, Z., Kong, L., and He, J. (2024). AgentBoard: An Analytical Evaluation Board of Multi-turn LLM Agents.

\bibitem{Ossowski2024}
Ossowski, T., Chen, J., Maqbool, D., Cai, Z., Bradshaw, T., and Hu, J. (2024). COMMA: A Communicative Multimodal Multi-Agent Benchmark.

\bibitem{Tan2024}
Tan, S., Zhuang, S., Montgomery, K., Tang, W. Y., Cuadron, A., Wang, C., Popa, R. A., and Stoica, I. (2024). JudgeBench: A Benchmark for Evaluating LLM-based Judges.

\bibitem{WorFBench}
Qiao, S., Fang, R., Qiu, Z., Wang, X., Zhang, N., Jiang, Y., Xie, P., Huang, F., and Chen, H. (2024). Benchmarking Agentic Workflow Generation.

\bibitem{Reflexion}
Shinn, N., Cassano, F., Gopinath, A., Narasimhan, K., and Yao, S. (2024). Reflexion: Language agents with verbal reinforcement learning. In \emph{Advances in Neural Information Processing Systems (NeurIPS)}, 36.

\bibitem{AutoGenStudio}
Dibia, V., Chen, J., Bansal, G., Syed, S., Fourney, A., Zhu, E., Wang, C., and Amershi, S. (2024). AutoGen Studio: A No-Code Developer Tool for Building and Debugging Multi-Agent Systems.

\bibitem{MultiAgentInAgentScope}
Pan, X., Gao, D., Xie, Y., Chen, Y., Wei, Z., Li, Y., Ding, B., Wen, J., and Zhou, J. (2024). Very Large-Scale Multi-Agent Simulation in AgentScope.


\end{thebibliography}

\newpage

\appendix

\section{Background}
\paragraph{Guardrails for Scaled Production Generative Systems with Human as Input}
For generative systems, particularly those involving LLMs, it is essential to implement guardrails to ensure safety, efficiency, and accountability, especially when human input is involved. These guardrails address potential risks such as constraint hallucinations, reduced interpretability, scaling issues, and security vulnerabilities. The Agent-as-a-Judge framework \cite{AgentAsAJudge} addresses the challenge of constraint hallucinations by enabling agents to evaluate each other throughout the generation process. This self-assessment mechanism fosters continuous feedback, promoting more dynamic and scalable improvements in agent performance and ensuring that the system adheres to the intended constraints during workflow generation.

\vspace{1em}

To enhance the interpretability of multi-agent systems, Triantafyllou et al. (2024) \cite{Triantafyllou2024} introduces a causal explanation model that attributes counterfactual effects to agents and state variables. This model aids in understanding how decisions are made within the system, increasing transparency and trustworthiness, which are essential when human involvement is integrated into the process.

\vspace{1em}

When scaling multi-agent LLM systems, frameworks such as Optima \cite{Chen2024} optimize communication and efficiency by refining task selection and training processes, achieving significant performance gains — up to 2.8x — and reducing token usage by 10 percent in information-heavy tasks. This ensures that agents can collaborate effectively without overwhelming the system resources.

\vspace{1em}
Security is also a critical consideration in scaled systems. Prompt Infection \cite{Lee2024} identifies vulnerabilities in multi-agent LLM environments, where malicious prompts can propagate like viruses, posing risks such as data theft and misinformation. The introduction of LLM Tagging mitigates this threat by marking potentially dangerous interactions, reinforcing the need for robust security measures in complex agent-based systems.

\vspace{1em}

In addition, studies such as Lermen et al. (2024) \cite{Lermen2024} highlight safety gaps, revealing how models such as Llama 3.1 can still perform harmful tasks despite safeguards. This issue is addressed by the Safe Agent Benchmark, which evaluates and improves the security of agentic systems, underscoring the need for more rigorous safety mechanisms to protect against unintended harmful actions.

\paragraph{Workflow Evaluation and Benchmark} 

Evaluating the performance of agents in workflow generation is critical for ensuring the effectiveness and scalability of these systems, yet current benchmarks often fail to capture the complexity and real-world applicability of generated workflows. Most existing benchmarks are results-driven, focusing primarily on whether the workflow produces the correct answer. For example, Zhuge et al. (2024) \cite{GPTSwarm} introduces a utility function that includes various cost dimensions, but the benchmark still largely measures the ability to generate the right answer, often in simple scenarios. Similarly, ChainBuddy \cite{ChainBuddy} parameterizes metrics such as the number and types of nodes in a workflow, but does not go far enough in capturing the full scope of workflow execution costs and real-world constraints.

\vspace{1em}

Current benchmarks generally fail to evaluate workflows based on the costs associated with their execution. To our knowledge, no existing framework thoroughly examines how efficiently and effectively a workflow is executed in practice, especially when considering resource consumption, time constraints, or the need for human intervention. This is a significant gap, as workflow execution involves more than just producing an answer, it is about doing so under specific, often complex conditions.

\vspace{1em}

Recent works of Ma et al. (2024) \cite{Ma2024} evaluate LLM agents in diverse, partially observable environments using a unified framework, but this framework still does not fully address the intricacies of workflow execution in real industrial use cases. Ossowski et al. (2024) \cite{Ossowski2024} evaluates multi-agent systems in collaborative settings where agents have unequal access to information, but the focus remains on communication and collaboration, not on the end-to-end execution of workflows. Tan et al. (2024)\cite{Tan2024} tests the reliability of LLM-based judges, but again the focus is on the agents’ assessment capabilities, not the execution of the workflows themselves.

\vspace{1em}

The ItineraryBench \cite{Wang2024} provides a valuable contribution by evaluating agents handling progressively complex interconnected tasks, but it primarily assesses planning and tool usage without incorporating the cost of execution or downstream impact. WorFBench \cite{WorFBench} evaluates LLM agents that generate workflows, emphasizes node similarities and logic but does not account for execution costs, limiting its practical application for real-world workflows.

\vspace{1em}

Given these limitations, our approach diverges by focusing on evaluating workflows within the context of real industry use cases. This involves a more holistic view of workflow execution, considering not just correctness but also efficiency, resource consumption, and the impact on downstream Tasks. By taking this direction, we aim to develop a comprehensive framework that evaluates workflows based on their execution and cost-efficiency in realistic, industry-relevant environments.

\paragraph{Reinforcement Learning for Workflow Generation and Optimization}
While we acknowledge recent advancements in leveraging reinforcement learning (RL) for task-oriented agents, we do not apply RL in this version of our workflow generation approach. This decision is driven by the complexity of workflow environments, where the action space is not well-defined or simple enough for traditional RL methods to be directly applicable. In complex workflows, actions are often nuanced and context-dependent, making it challenging to model the environment in a way that would allow for effective RL-based decision-making. Instead, we focus on approaches that better handle the intricate structures of real-world workflows. Reflexion \cite{Reflexion} proposes a framework for reinforcement learning for language agents that replaces traditional weight updates with linguistic feedback. Agents reflect on task feedback, storing it in episodic memory to improve future decision-making. From Words to Actions \cite{FromWordsToActions}, mentioned before, presents a hierarchical RL model for LLMs, where a Planner generates sub-goals for the Actor to execute. The Planner uses Bayesian aggregated imitation learning (BAIL) and an $\epsilon$-greedy exploration strategy to reduce regret. This approach enables effective decision-making in partially observable environments and supports multi-agent coordination through the Planner acting as a world model.

\paragraph{Workflow Manual Builder systems}

AutoGen Studio \cite{AutoGenStudio} and ChainBuddy \cite{ChainBuddy} are systems designed to facilitate the creation and evaluation of multi-agent workflows. AutoGen Studio, built on the AutoGen framework, allows users to prototype, debug, and evaluate workflows using a no-code interface. It enables the specification of LLM-enabled agents through a declarative JSON-based format and offers tools for interactive evaluation and debugging. ChainBuddy, part of the ChainForge platform, assists users in generating and evaluating LLM pipelines for specific tasks. It simplifies the process of setting up and evaluating workflows, aiming to reduce the ``blank page" problem. AgentScope \cite{MultiAgentInAgentScope} introduces an easy-to-use configurable tool and an automatic background generation pipeline, enabling users to efficiently create diverse agents with detailed settings, alongside its actor-based distributed mechanism for scalable and efficient workflow execution. These systems heavily rely on user interaction to define tasks but are limited in their approach to task structuring and definitions, which we find inadequate for the complexities of our use cases.

\paragraph{Opus AI Knowledge Work Platform}

Opus allows organizations to generate, optimize and execute Workflows incorporating human expert reviews across each Workflow execution to deliver supervised automation. Users can design their Workflows in a powerful Workflow builder, browse and customize Workflows from the Opus Workflow library. Users can also provide Workflow description, sample Input / Output files or existing outsourcing contracts for Opus to perform inference across the workflow generation system, powered by the Opus LWM and WKG, to suggest optimal Workflows based on specific process needs and context. Individual jobs can then be executed via Opus system integrations and the Opus job management interface to run Opus Workflows in daily operations.

\section{Medical Coding Reference Workflow}
Figure 3 provides a high-level overview of the reference Workflow used in our experiments, offering a simplified semantic description of the tasks involved in the CPT\textsuperscript{\textregistered} Evaluation / Management (E/M) Coding process following the American Medical Association's 2021 guidelines. CPT\textsuperscript{\textregistered} is a registered trademark of the American Medical Association. The codes and descriptions used in this publication are for reference purposes only and are not intended for billing or reimbursement.

\begin{figure}[H]
\centering
\includegraphics[width=\linewidth]{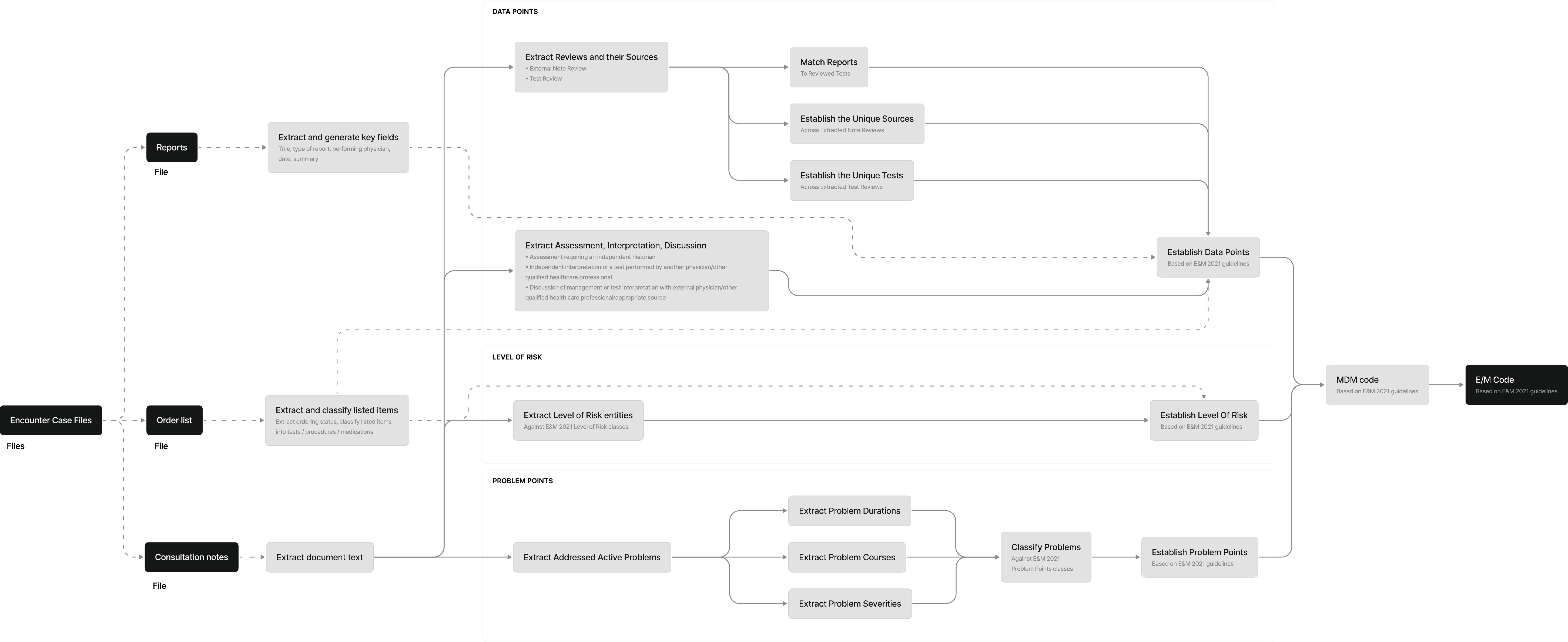}
\caption{Medical Coding Opus Reference Workflow — simplified view}
\label{fig:med_coding_workflow}
\end{figure}

\newpage

\section{Results}

\begin{figure}[H]
\begin{center}
    \begin{tabular}{cc}
        {\tiny
        \includegraphics[width=0.35\textwidth]{images/results/appliedai-opus-1alpha-large.eps}
        } &
        {\tiny
        \includegraphics[width=0.35\textwidth]{images/results/appliedai-opus-1alpha-small.eps} 
        }
        \\
        \small{Opus Alpha 1 Large} & \small{Opus Alpha 1 Small} \\
        {\tiny
        \includegraphics[width=0.35\textwidth]{images/results/openai-o1-preview-2024-09-12.eps}
        } & 
        {\tiny
        \includegraphics[width=0.35\textwidth]{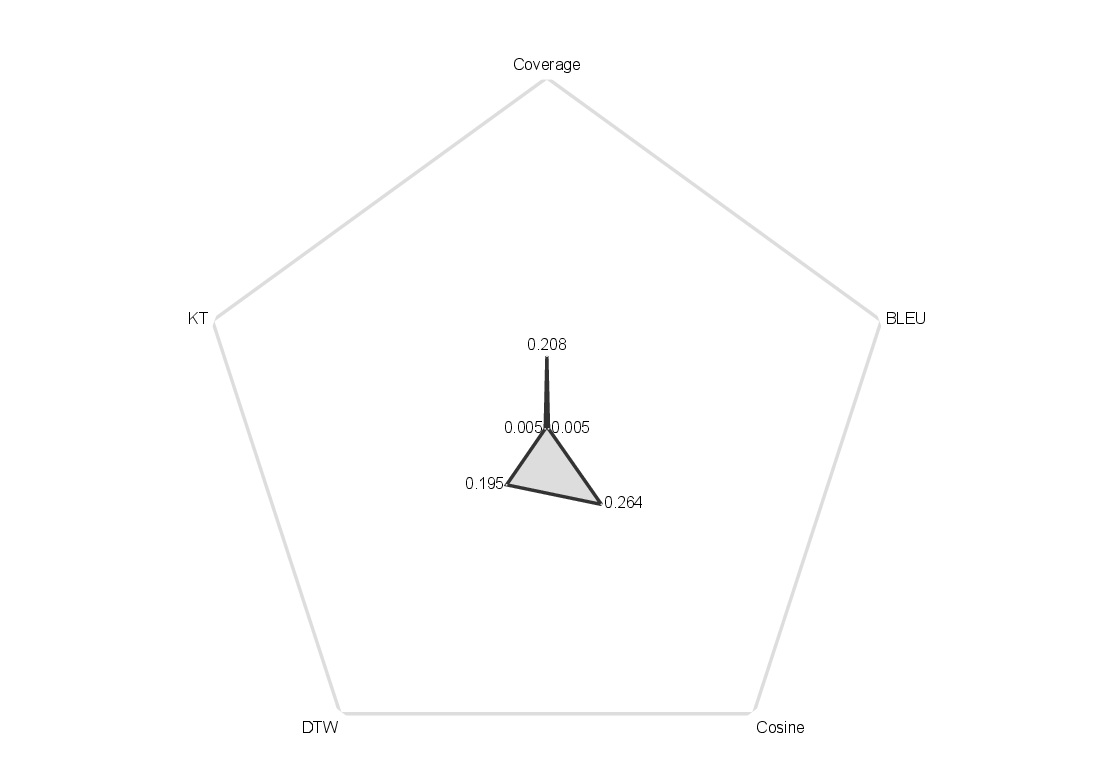} 
        } \\
        \small{OpenAI o1-preview} & \small{OpenAI 4o} \\
        {\tiny 
        \includegraphics[width=0.35\textwidth]{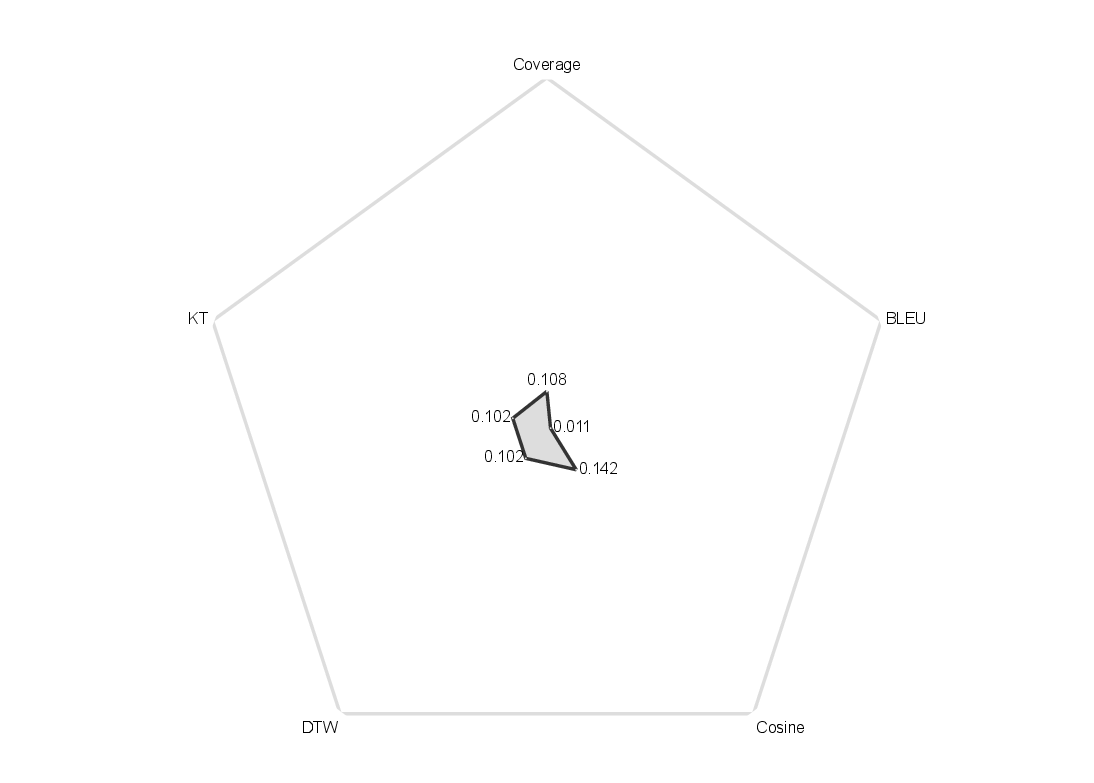} 
        } &
        {\tiny
        \includegraphics[width=0.35\textwidth]{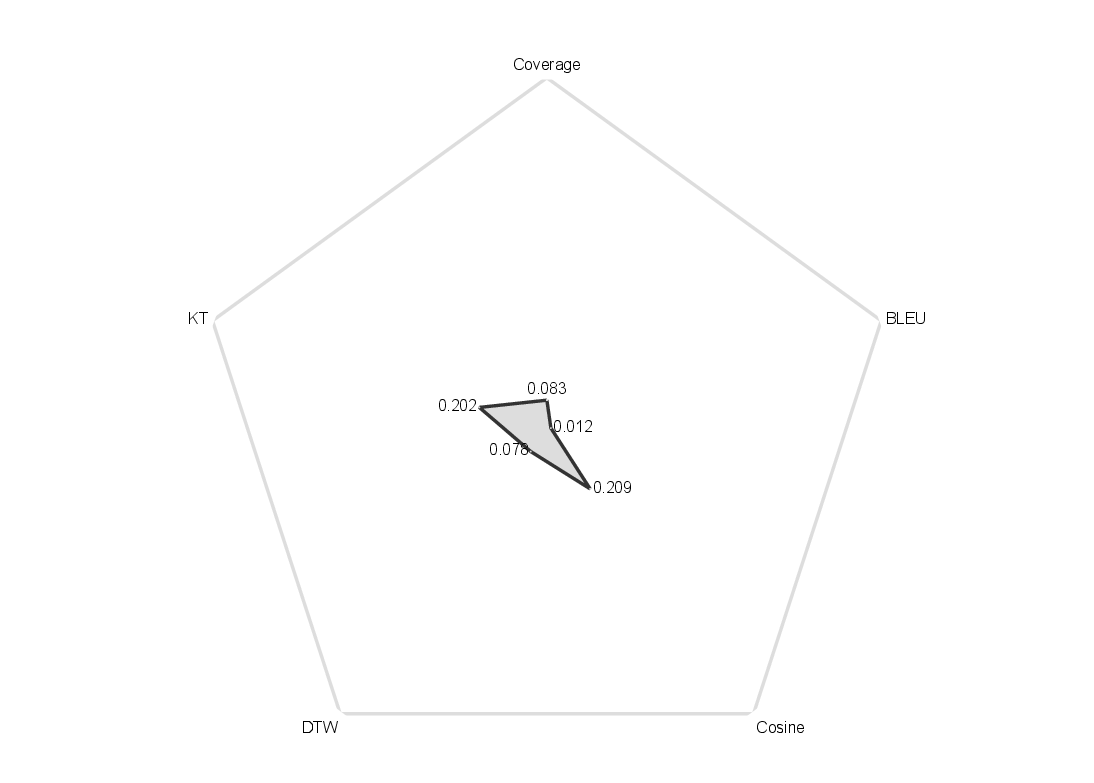} 
        } \\
        \small{Gemini Pro} & \small{Gemini Flash} \\
    \end{tabular}
    {\tiny \includegraphics[width=0.35\textwidth]{images/results/anthropic-claude-3.5-sonnet.eps} 
    } \\ 
    \small{Anthropic Claude 3.5 Sonnet}
\end{center}
    \caption{State-of-the-art LLMs performance across the metrics on the Medical Coding use case.}
    \label{fig:allresults}
\end{figure}

\section{Acronyms and Definitions} 

\begin{table}[H]
\centering
\label{tab:acronyms_definitions}
\renewcommand{\arraystretch}{1.25}
\begin{tabular}{>{\centering\arraybackslash}m{1.5cm}>{\raggedright\arraybackslash}m{14cm}}
\hline
\textbf{Acronym} & \multicolumn{1}{c}{\textbf{Definition}} \\ \hline \hline

\textbf{$x$} & \textbf{Instruction}: model inference, executable code (logical function, tool, etc.) or a human expert review.  \\ \hline
\textbf{t} & \textbf{Task}: executable sequence of Instructions. \\ \hline
\textbf{w} & \textbf{Workflow}: Directed Acyclic Graph whose nodes are Tasks and edges are the flow of execution.\\ \hline
\textbf{$\Gamma$} & \textbf{Encoded Intention}: vector representation of the Client's Input, Output and Workflow Process Context.\\\hline
\textbf{LWM} & \textbf{Large Work Model}: fine-tuned Large Language Model that generates Workflows as sequences of Tasks.\\ \hline
\textbf{WKG} & \textbf{Work Knowledge Graph}: graph storing semantic and historical implementations of Workflows from many industries. \\ \hline
\textbf{SWKG} & \textbf{Sub-Work Knowledge Graph}: sub-graph of the WKG relevant to $\Gamma$. \\ \hline
\textbf{WFG} & \textbf{Workflow Graph}: Directed Graph comprised of LWM generated Workflows with WKG edges and nodes. \\ \hline

\textbf{BPO} & \textbf{Business Process Outsourcing}: outsourcing of specific business processes to a third-party service provider. \\ \hline
\textbf{E/M} & \textbf{CPT\textsuperscript{\textregistered} Evaluation / Management Coding}: a set of codes developed by the American Medical Association (AMA) used by healthcare providers to document and bill for patient encounters that involve evaluating and managing a patient's health. The codes evaluate the risk level, medical data and problem complexities of a medical encounter. CPT\textsuperscript{\textregistered} is a registered trademark of the American Medical Association. The codes and descriptions used in this publication are for reference purposes only and are not intended for billing or reimbursement. \\ \hline

\end{tabular}
\end{table}

\end{document}